\documentclass[conference]{IEEEtran}
\IEEEoverridecommandlockouts                            
 
\usepackage{url}
\usepackage[bookmarks=false]{hyperref}
\usepackage{graphicx} 
\usepackage{pbox}
\usepackage{enumerate}
\usepackage{booktabs}
\usepackage{amsmath}
\usepackage{algorithm2e}
\usepackage{algorithmic}
\usepackage{listings}
\usepackage{color}
\usepackage{multirow}
\usepackage{tikz}
\usetikzlibrary{positioning}

\newcommand\copyrighttext{%
  \footnotesize \textcopyright 2019 IEEE. Personal use of this material is permitted. Permission from IEEE must be obtained for all other uses, in any current or future media, including reprinting/republishing this material for advertising or promotional purposes, creating new collective works, for resale or redistribution to servers or lists, or reuse of any copyrighted component of this work in other works.
  DOI: \href{https://doi.org/10.1109/ISSREW.2019.00102}{10.1109/ISSREW.2019.00102}}
\newcommand\copyrightnotice{%
\begin{tikzpicture}[remember picture,overlay]
\node[anchor=south,yshift=10pt] at (current page.south) {\fbox{\parbox{\dimexpr\textwidth-\fboxsep-\fboxrule\relax}{\copyrighttext}}};
\end{tikzpicture}%
}

\definecolor{dkgreen}{rgb}{0,0.6,0}
\definecolor{gray}{rgb}{0.5,0.5,0.5}
\definecolor{mauve}{rgb}{0.58,0,0.82}

\title{\LARGE \bf
Automatic Cause Detection of Performance Problems in Web Applications
}

\author{\IEEEauthorblockN{Quentin Fournier\IEEEauthorrefmark{1},
Naser Ezzati-jivan\IEEEauthorrefmark{2}, Daniel Aloise\IEEEauthorrefmark{3}, and Michel R. Dagenais\IEEEauthorrefmark{4}}
\IEEEauthorblockA{Polytechnique Montreal, Quebec H3T 1J4\\
Email: \IEEEauthorrefmark{1}quentin.fournier@polymtl.ca,
\IEEEauthorrefmark{2}n.ezzati@polymtl.ca,
\IEEEauthorrefmark{3}daniel.aloise@polymtl.ca},
\IEEEauthorrefmark{4}michel.dagenais@polymtl.ca
}

\begin{document}

\maketitle
\thispagestyle{empty}
\pagestyle{empty}

%
\copyrightnotice

\begin{abstract}
The execution of similar units can be compared by their internal behaviors to determine the causes of their potential performance issues. For instance, by examining the internal behaviors of different fast or slow web requests more closely, and by clustering and comparing their internal executions, one can determine what causes some requests to run slowly or behave in unexpected ways. In this paper, we propose a method of extracting the internal behavior of web requests as well as introduce a pipeline that detects performance issues in web requests and provides insights into their root causes. First, low-level and fine-grained information regarding each request is gathered by tracing both the user space and the kernel space. Second, further information is extracted and fed into an outlier detector. Finally, these outliers are then clustered by their behavior, and each group is analyzed separately. Experiments revealed that this pipeline is indeed able to detect slow web requests and provide additional insights into their true root causes. Notably, we were able to identify a real PHP cache contention issue using the proposed approach.
\end{abstract} 


\section{Introduction}
\label{introduction}

While it is crucial to identify performance issues, they can also be extremely difficult to detect \cite{7469402} as they are rare and hard to reproduce. Additionally, detection techniques usually require specifying an unknown normal behavior.

Once a performance issue has been detected, developers often use debuggers and profilers to analyze the issue and identify its root cause. However, debuggers are rarely applicable as they operate by stopping the world -- the program -- which may, in fact, mask problems related to latency or race conditions. Profilers are also ineffective as they operate by averaging metrics, which hides outliers.

Execution tracing overcomes the drawbacks of debuggers and profilers by instrumenting the system at different levels and by collecting information at run time. It works by executing a macro that generates an event for each instance of tracepoints met during execution. Tracing permits low-level and fine-grained information to remain on the system, which aids in discovering the root cause of the problem. However, the size of the collected trace may be enormous in complex, modern systems.

As the size and complexity of trace data continue to increase, the need for automated analysis becomes equally as important. However, automating this process is complex. A system may perform slowly for numerous reasons: an improper configuration, a change in the environment, unusual and possibly malicious network traffic, a software bug, or simply an inefficient code modification. An expert is only able to dissociate the aforementioned performance issues by comparing the anomaly with normal executions.

Detecting whether a given request is normal or abnormal is a challenging task. Some may assume that only a request response time is relevant to examine and, therefore, believe it is a trivial task. Yet, response time is not the sole factor to be considered. For example, two requests with the same time duration can have different internal behaviors (e.g., they generate different numbers of page faults, or they behave differently when they access the disk). Although they have identical response time, the one with an abnormal behavior can foreshadow an impending problem. Therefore, a general method of grouping requests based on their internal behavior is necessary. This grouping will be used for post-mortem root cause analysis.

This observation brings two research questions: (1) Can anomalies be efficiently and automatically detected from trace data? (2) Can the root cause of the anomaly be identified by comparing its internal run time behavior to that of a normal one? We restrict the scope of this paper to the requests, that we consider to be any task separated by a specific start and end event. Valid examples are web requests, database requests, and multiple calls to the same function in any application. In particular, we would like to discover which internal action or behavior causes a web request to run slowly.


Our main contributions are: (1) Tracing different levels of requests executions and extracting their internal behavior by following all notable threads along the critical path, rather than merely a single thread. (2) A proposed pipeline to detect outliers, and cluster them according to their run time behavior, which will enable comparison and evaluation. We show that the extracted features are especially useful for detecting outliers.

The remaining part of the paper is organized as follows. Section \ref{related_work} presents the related work. Section \ref{proposed_approach} introduces our proposed approach to extract meaningful features, detect anomalies, and subsequently identify their root cause. Section \ref{results} details our results and meticulously analyzes the complexity of the proposed approach. Finally, Section \ref{conclusion} concludes this work.

\section{Related Work}
\label{related_work}

Dynamic analysis through execution tracing is used to analyze software behavior \cite{ezzatisurvey,systematicabstractionsurvey}. The analysis of kernel tracing data is studied in \cite{giraldeau}, while the processing of system call traces are studied in \cite{ezzati,wahab-correlation}. The visibility of most of the existing works is, however, limited to only a single layer of the system, while one would typically need visibility of multiple layers (e.g., application, system calls, network, etc.) in order to completely analyze internal software behaviors \cite{Murtaza2013}. In this paper, we analyze traces collected from both the application and kernel layers, further allowing for more effective insights into the software run time behavior.

As the complexity and variety of modern systems increase, so does the need to automatize their analysis. Recent progress in machine learning (ML) contributes interesting results for automating diagnosis tasks like intrusion and malware detection \cite{Kim2016} or code correlation and optimization \cite{8357388}. These techniques can be used to automatically detect changes in code \cite{abder1}, configuration, environment, run time behavior, and performance \cite{Yuan2011,7469402}. Unfortunately, the variety of tracing formats, the large size of run time data, as well as their unstructured nature present additional challenges in both the data modeling and processing steps which require special care. 

In the literature, kernel traces are represented in a variety of methods, but two primary approaches emerge. The first one preserves the ordering and the temporal information. These representations are sequences of system call names, learned system call embedding \cite{Kim2016}, and kernel states \cite{Murtaza2013}. The second approach aggregates the sequences, trading the temporal information for a more compact representation. The main example of this approach is called bag-of-word, also known as \textit{system call counts vector} \cite{DBLP:journals/corr/DymshitsMT17}, \textit{frequency counts of system call names} \cite{1495972} or \textit{bag of system calls} \cite{2c4c5c42cfef4dcbb0f268819914a44d}. N-gram is a technique which lies in between the two approaches as it preserves only a short and local ordering. This method has been extensively used for anomaly detection and intrusion detection as they are more expressive \cite{1495972, 7272922, Wressnegger:2013:CLN:2517312.2517316}. This work falls into the second approach, with the exception that duration is used in addition to counts.

Clustering anomalies based on their behavior helps to better understand, predict, and maintain them. Ideally, one would prefer to cluster anomalies based on their underlying and unknown root cause. A practical approach is to cluster them based on system resources consumption behaviors, thus grouping the faulty behaviors based on the extent to which they compete over resources to serve user requests. In a recent work, Nemati et al.\cite{hani:19} extracted high-level features from low-level traces of virtual machines and grouped them using two-stage k-means. Their method provided insights into the different behaviors and potential anomalies. Their work, however, relies on data collected from only one layer, which restricts the overall visibility. 

This paper addresses the above-mentioned limitation by proposing an approach based on a multi-level data collection policy to ensure maximum visibility while undertaking the root cause analysis. Collecting data from different levels provides a more complete view of the system dynamics (application, operating system, network, etc.) and ensures that a broad range of anomaly can be detected. Indeed, although the application may be bug-free, a contention for resources between instances of the same application can be the source of inexplicable latencies. Such a complication is only visible from the operating system level.

Furthermore, our data collection process exceeds the simple tracing of single threads, which is the case in most of the aforementioned research works. Our method follows a request across all interacting threads, and possibly across different machines, to ensure that all required levels of details are collected.

Although using off-the-shelf, standard, unsupervised techniques, this new approach, with a pipeline to detect, cluster and analyze anomalies in requests, has not been proposed earlier, to the best of our knowledge.
\section{Proposed Approach}
\label{proposed_approach}

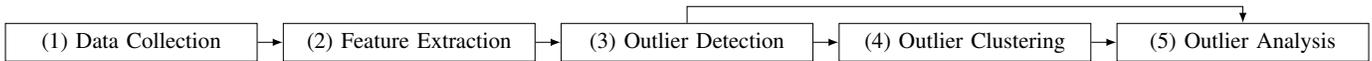
\begin{figure*}[tp]
\centering
\resizebox{\textwidth}{!}{\begin{tikzpicture}
    [block/.style={draw,rectangle,minimum width=100pt,minimum height=15pt, text centered,node distance=10pt,font=\small}]
    \node[block] (f1) {(1) Data Collection};
    \node[block,right=of f1] (f2) {(2) Feature Extraction};
    \node[block,right=of f2] (f3) {(3) Outlier Detection};
    \node[block,right=of f3] (f4) {(4) Outlier Clustering};
    \node[block,right=of f4] (f5) {(5) Outlier Analysis};
    \draw[->, >=latex] (f1) -- (f2);
    \draw[->, >=latex] (f2) -- (f3);
    \draw[->, >=latex] (f3) -- (f4);
    \draw[->, >=latex] (f3) |- ++(7.5,0.5) -| (f5);
    \draw[->, >=latex] (f4) -- (f5);
\end{tikzpicture}}
  \caption{The proposed pipeline for anomaly detection.}
  \label{fig:pipeline}
\end{figure*}

\subsection{Data Collection}
Gathering applicable data is the first step in the proposed pipeline, as summarized in Figure \ref{fig:pipeline}. To that extent, we used the Linux Trace Toolkit Next Generation (LTTng) \cite{desnoyers2008lttng}, a low-overhead open-source tracing tool that collects data from several layers. Data was acquired from the user space level, to distinguish the start and end of each request, as well as from the kernel level, to collect information regarding what actually occurs within the system during each request.

User space tracers typically work by instrumenting the source code (i.e., the web application source code) or the language core (i.e., the PHP core). In this work, the latter is instrumented because it adds a new tracing extension that inserts tracing macros in different entry points (function calls, request entry and exit, etc.).  This ensures that users do not need to change their own source code. Each time the execution in the language core reaches a tracepoint, the associated macro is executed, generating an event that is stored in the corresponding CPU buffer. Table \ref{phpstack} shows all the events that the PHP tracer is able to collect. This PHP extension is open-source and publicly available\footnote{https://github.com/naser/LTTng-enabled-PHP}.

\begin{table}[htb]
\centering
\caption{Tracepoints in the developed PHP tracing extension}
\label{phpstack}
\begin{tabular}{ p{1.8cm}  l  p{4cm}}

\\[-0.7em]\textbf{Trace Event}         & \textbf{Description}                                     \\ \hline
\\[-0.5em]request\_start   & Fires when a new PHP request is arrived. \\ 
\\[-0.5em]request\_exit   & Fires when the handing of a PHP request is completed.                                         \\ 
\\[-0.5em]function\_start   & Fires when a function is called.                             \\ 
\\[-0.5em]function\_exit   & Fires when a function exits. \\ \hline
\end{tabular}
\end{table}

Kernel tracers, on the other hand, usually work by instrumenting the different parts of the operating system. Luckily, the Linux kernel is already instrumented and has over two hundred incorporated tracepoints. The LTTng tracer attaches to these tracepoints and gathers data on system calls, processes, file systems, disk accesses, memory accesses, network layers, interrupts, timers, and other relevant areas.

\subsection{Feature Extraction}
The raw collected trace data is initially in a semi-structured, multi-dimensional, and multi-level format which needs to be projected in a lower-dimensional vector space before being analyzed. Indeed, most clustering algorithms accept only fixed-size arrays and are sensitive to the quality and compactness of the representation\footnote{See the curse of dimensionality for an example of a bad representation.}.

First, the trace must be divided into individual requests. The collected trace contains events from multiple web requests concurrently received by the server. Only two events from user space, namely \texttt{request\_start} and \texttt{request\_exit}, are required to correctly identify the boundaries of each request, received by a web server like Apache, and are passed on to the PHP Engine. Intuitively, \texttt{request\_start} is fired when the PHP Engine accepts a request and \texttt{request\_exit} when its handling is completed.

\begin{figure*}[!ht]
 \centering
    \includegraphics[width=\linewidth]{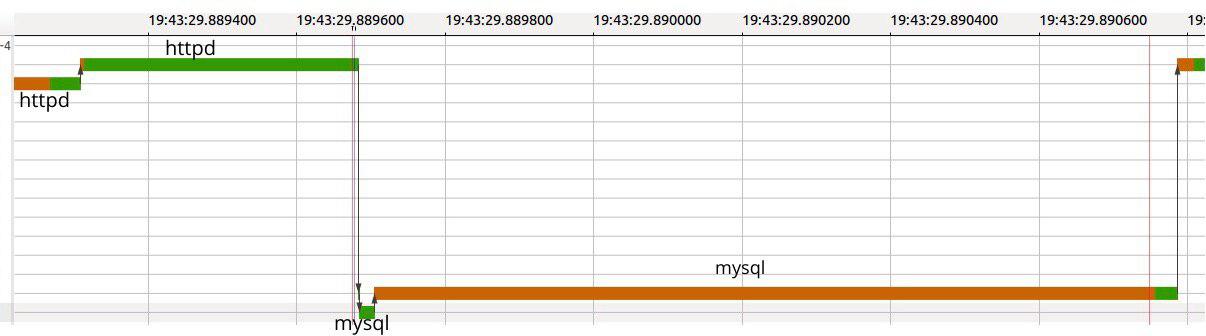}
    \caption{Critical path of a single web request spanning four threads.}
    \label{criticalpath}
\end{figure*}

The next step is to collect detailed metrics from the kernel trace data to create a feature set for each request. A request can be handled by a single thread or, as it is the case with the majority of existing servers, by multiple threads. For instance, the latter can be the collaboration of a web server (e.g., Apache web Server or Nginx), a web application (e.g., PHP or Node), and a database (e.g., MySQL or SQL Server). Our analysis is based on features extracted from the critical path of each request, which spans across different threads. Figure \ref{criticalpath} shows the critical path of a typical web request. The request time is broken into five intervals across four interacting threads, showing the detailed contribution of each thread to the total response time. 

To discover the critical path of each request, we rely on the algorithm introduced by Giraldeau et al. \cite{fgiraldeau}. They proposed an algorithm that would extract the active execution path across threads which possibly run on distant machines. Their method extracts various execution states for each thread, including running, interrupt handling, waiting for disk, waiting for network, waiting for timer, and waiting for another task. The algorithm first builds an execution graph showing all interactions between threads. Then the critical path -- the minimal active path -- is extracted by replacing recursively the waiting edges of a thread by the edges of the waking thread, from within the execution graph. In the final graph, for which an example is shown in Figure \ref{criticalpath}, vertical edges denote the waker/wakee relationships between the threads while the horizontal edges represent the different execution states within a thread.

The method proposed by Giraldeau et al. \cite{fgiraldeau} was adapted to retrieve metrics regarding the different execution states that contribute to the response time of each request. Notably, the following execution states are considered:
\begin{itemize}
\item Running in system call mode (RS)
\item Running in user mode (RU)
\item Blocked for disk I/O (BD)
\item Blocked for network (BN)
\item Blocked for CPU (preempted) (BP)
\item Blocked for another task (BT)
\item Blocked for futex (BF)
\item Blocked for interrupt (BI)
\item Blocked for timer (BS)
\item Total request time (TT)
\end{itemize}

Table \ref{tab:feature_set} shows the four different feature sets that were created for each request based on the above execution states.
\begin{table}[!ht]
\centering
\caption{Request feature sets used for clustering}
\begin{tabular}{ll}
Feature set & Example \\ \hline
System calls sequence & \texttt{open, seek, read, close...} \\
Execution states sequence & \texttt{RU, RS, BD, BT, BP...} \\
Execution states count & 15, 0, 1, 23, 110... \\
Execution states total duration & 0.12, 0.0, 0.01, 0.85...\\ \hline
\end{tabular}
\label{tab:feature_set}
\end{table}



\subsection{Outlier Detection}

In order to automatically detect performance issues, one method includes first detecting outliers -- out-of-distribution data points -- then applying domain-specific tools to discover the root cause of the anomaly. However, the detection of outliers is highly dependent upon the data representation. In this paper, we propose a framework to cluster abnormal requests based on the features previously extracted. This section corresponds to the pipeline third step.

Clustering algorithms minimize the inter-point distance within a cluster while simultaneously maximizing the distance between clusters. However, choosing the \textit{right} number of clusters is not a trivial task. Moreover, because the points of interest are abnormal, the method needs to provide an efficient clustering of outliers.

Density-Based Spatial Clustering of Applications with Noise (DBSCAN) \cite{Ester:1996:DAD:3001460.3001507} is a well-known clustering algorithm which does not require the number of clusters to be specified. Instead, it accepts two parameters: $\epsilon$ the maximum distance for two points to be considered neighbors and $m$ the minimum number of points required in order to create a cluster. DBSCAN works by first selecting a point $X$ at random. If $X$ has at least $m$ neighbors, then a cluster is created with $X$ and all of its neighbors. The cluster is expanded by then considering the neighbors of $X$. If $X$ has less than $m$ neighbors and is not part of a cluster, then it is considered an outlier. The process continues until all points are considered.

DBSCAN presents three key advantages: (1) the number of clusters is implicit, (2) an arbitrary shape of clusters are created, and (3) outliers are automatically separated. However, DBSCAN is sensible to its two parameters -- $\epsilon$ and $m$. Those parameters can be set manually or estimated with a random search. Furthermore, DBSCAN can only be applied to multi-dimensional data of a fixed size. Thus, the variable-length sequences of system calls and features must be converted to a fixed size array.

One approach to representing a variable-length sequence with a fixed size vector is to count the number of occurrences of each event. This technique is called "bag-of-word" (BoW) and has been extensively used in natural language processing. One may argue that events that appear across all executions carry little information. The term "frequency-inverse document frequency" (TF-IDF) is a popular technique to measure bag-of-word according to the importance of each event. More formally:
\begin{equation}
    \text{tf-idf}_{i,j}=\text{tf}_{i,j}\times\text{log}\left(\frac{|D|}{|d_j:t_i\in d_j|} \right)
\end{equation}
where $\text{tf}_{i,j}$ is the frequency of the $i$-th event in the $j$-th sequence, $|D|$ is the total number of sequence, and $|d_j:t_i\in d_j|$ is the number of sequence containing the $i$-th event.

Specific methods such as Markov Models have been developed to detect outliers within sequences. However, those methods assume that events are regularly generated, (i.e., there is a fixed duration between events). This crucial assumption does not hold true in the context of traces. Moreover, those techniques are computationally expensive. For those reasons, they are not included in the proposed framework.

\subsection{Outlier Clustering and Analysis}

Outliers are rare by definition, yet they may be numerous enough to require clustering before they can be analyzed. This corresponds to the optional fourth step in our pipeline.

The objective is to separate different types of abnormal behavior. To that extent, feature counts are more informative than feature duration. Indeed, a slow request has either a high count of non-blocking events or a small count of blocking ones. However, consider that a fast request may also have a high number of non-blocking system calls like \texttt{fnctl}\footnote{\texttt{fnctl} is the system calls that manipulate a file descriptor. For more information, see: \url{http://man7.org/linux/man-pages/man2/fcntl.2.html}}, surely indicating that something is behaving incorrectly.

K-means is a simple yet efficient clustering algorithm that has been widely used because of its ability to find tight spherical clusters. However, k-means is a distance-based method which means it is sensitive to data scale. To mitigate this effect, the feature counts have been standardized. The elbow method \cite{Skiena:2017:DSD:3152685} was applied in order to find the appropriate number of clusters. This is an ambiguous heuristic stating that one should choose the number of clusters, as long as it does not substantially decrease the inertia.

Once outliers have been clustered, each group is separately analyzed. This is the last step in our pipeline. As an example of a typical analysis, statistics about requests and average n-grams were computed independently, for each cluster of outliers as well as for non-outlier requests. Finally, we were able to link those results with real anomalies, confirming that the proposed pipeline is indeed able to detect outliers, group them appropriately, and bring forward useful information regarding the root cause of the anomaly.

\section{Results}
\label{results}

The complete project including the code, data, parameters, and results is available on GitHub\footnote{\url{https://github.com/qfournier/request_analysis}}.

As a proof of concept, around 50,000 requests were generated using the Apache benchmark suite\footnote{\url{https://httpd.apache.org/docs/2.4/fr/programs/ab.html}}. The experiment was performed using anywhere from 1 to 1,000 clients. In the PHP tracing module, only the two tracepoints that collect the start and end of each request were enabled. This gave a window for which we collected detailed kernel events.

\subsection{Outlier Detection}

Once requests were generated and features extracted, DBSCAN was applied to the different representations discussed in section \ref{proposed_approach}. The parameters $\epsilon$ and $m$ were set manually for each experiment.

The outlier detection was first evaluated qualitatively by comparing the distribution of durations for outlying and non-outlying requests. One would expect little overlap between the two groups as we know that slow requests are abnormal, hence outliers, and that fast ones are largely normal.

\begin{figure*}[tp]
\centering
\includegraphics[width=0.32\textwidth]{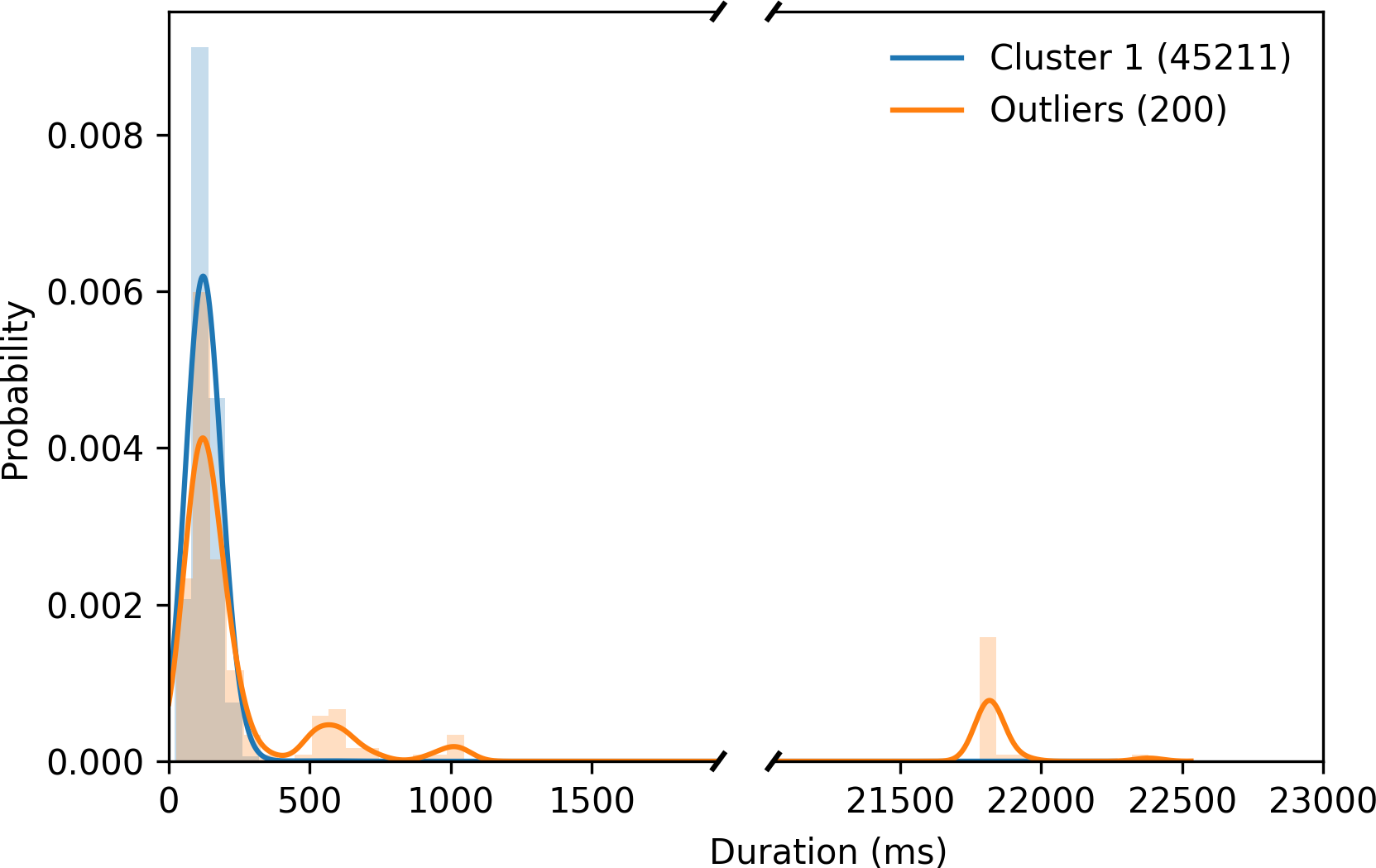}\hfill
\includegraphics[width=0.32\textwidth]{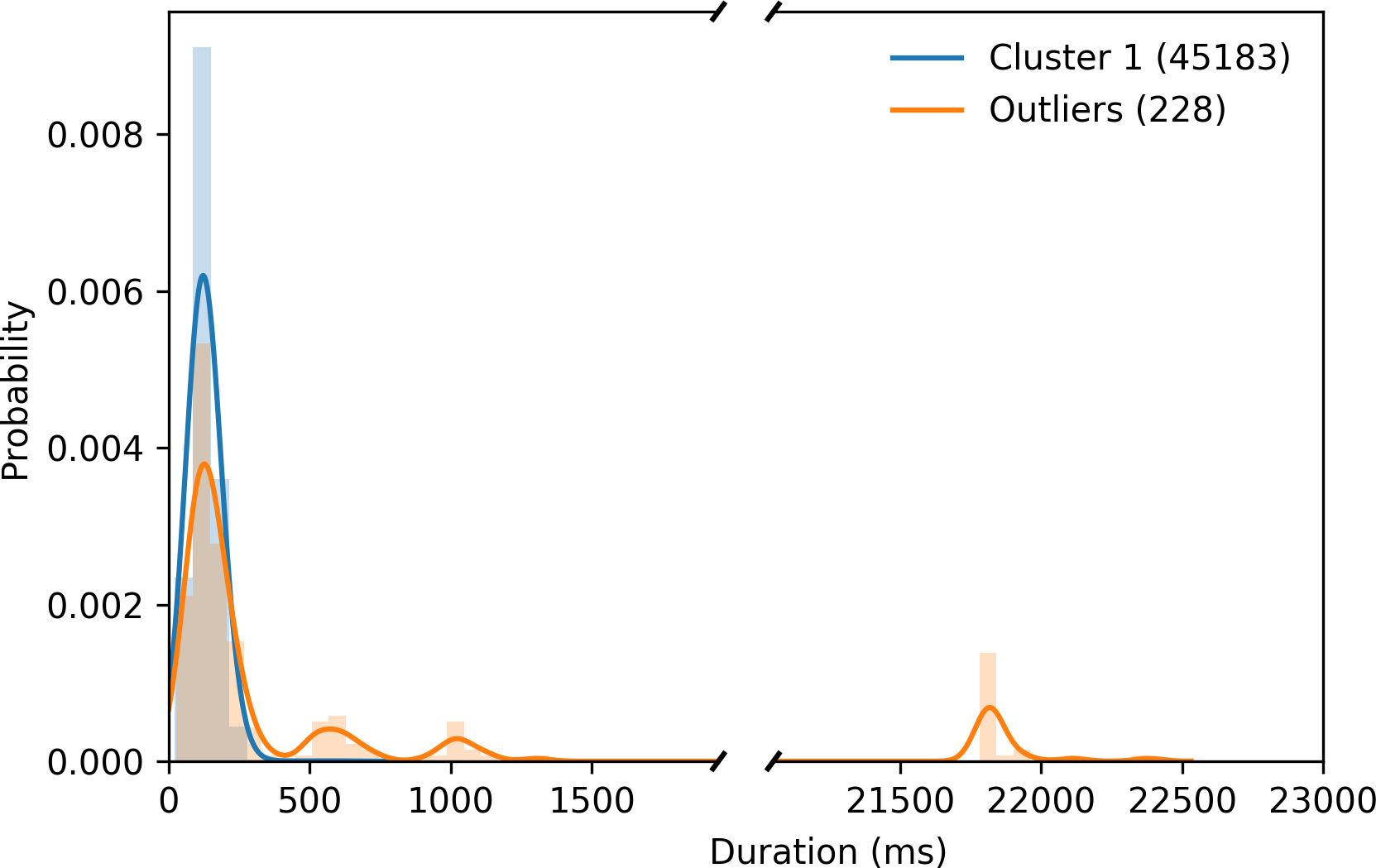}\hfill
\includegraphics[width=0.32\textwidth]{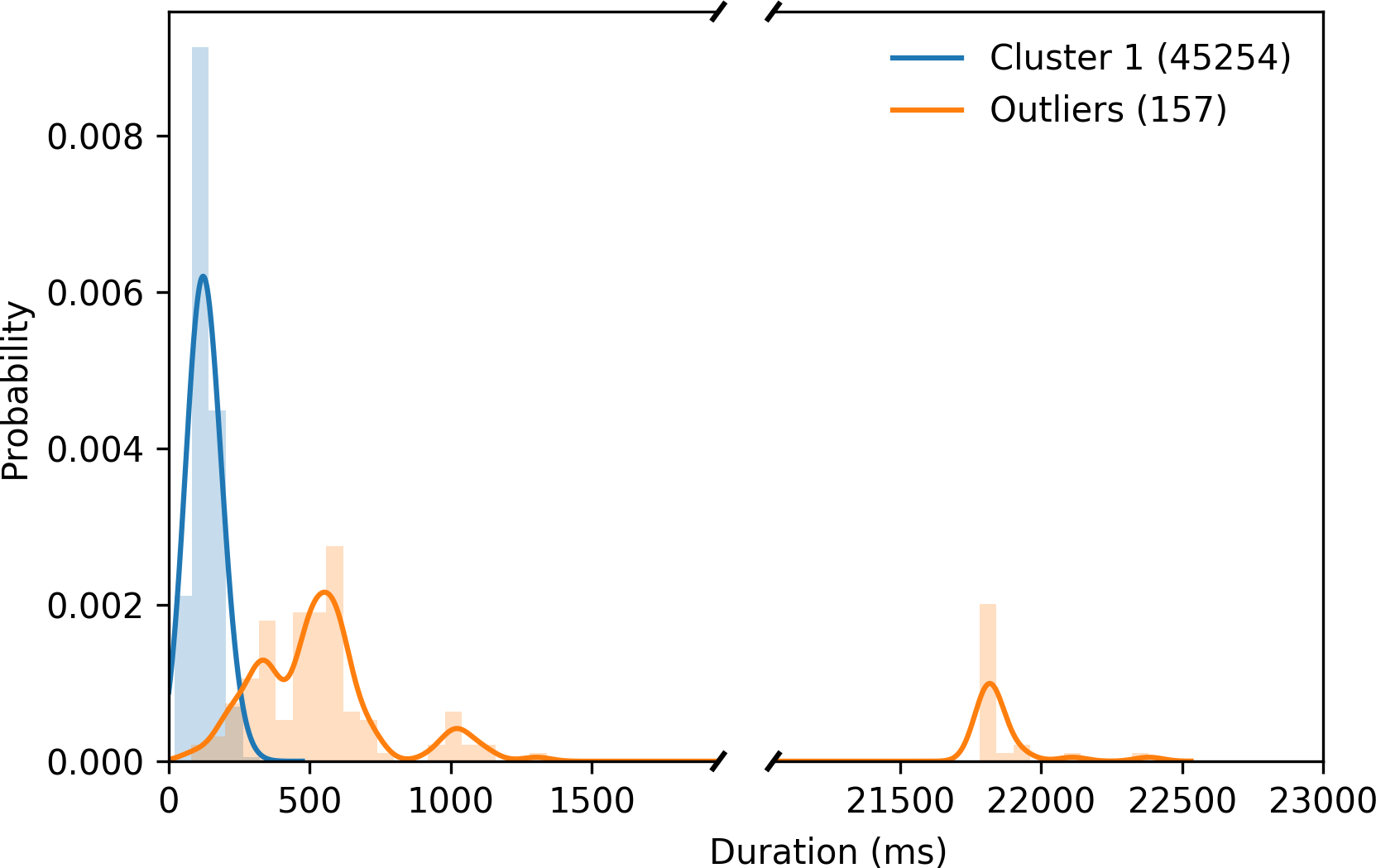}
\caption{Distribution of the duration of outlying and non-outlying requests. DBSCAN was applied on three different requests representations: (left) system call counts, (middle) feature counts, and (right) feature duration. The size of each cluster is specified between parenthesis in the legend.}
\label{fig:distribution}
\end{figure*}

System call counts and feature counts did not provide a clear separation of the two distributions (Fig. \ref{fig:distribution}). TF-IDF did not yield better results, because some events are so common that their weight is almost null. Only the feature duration was able to provide a clear separation, with little overlap between the outliers duration distribution and the non-outliers.

The outlier detection was then evaluated quantitatively by looking at two statistics: 
\begin{enumerate}
    \item The median duration of outlying requests. The median is preferred to the mean as it is less sensitive to a few miss-classified normal requests. One would expect the median duration of outliers to be significantly higher than for the whole data set.
    \item The probability of a detected outlying request having a duration above 200, 250, and 300 milliseconds. Those values are possible thresholds to detect slow requests, also known as anomalies. One would expect the probability to be high, but not equal to 100\% as some fast requests could also have an abnormal behavior.
\end{enumerate}

\begin{table}[htp]
\centering
\caption{outlying requests' statistics.}
\begin{tabular}{lrr|rrr}
    & \multicolumn{2}{c|}{System calls} & \multicolumn{3}{c}{Features} \\
    & BoW & TF-IDF & BoW & TF-IDF & Duration \\ \hline
    Number of outliers & 200 & 362 & 228 & 236 & 157 \\
    Median duration & 147 & 115 & 168 & 155 & \textbf{554}\\           
    $P(d>200)$& 0.36 & 0.163 & 0.408 & 0.390 &\textbf{0.968} \\
    $P(d>250)$& 0.28 & 0.108 & 0.316 & 0.305 & \textbf{0.924} \\
    $P(d>300)$& 0.26 & 0.940 & 0.276 & 0.267 & \textbf{0.892} \\\hline
\end{tabular}
\label{tab:outlier_statistics}
\end{table}

DBSCAN yielded an outlier detection rate lower than one percent for each representation, which is reasonable (Table \ref{tab:outlier_statistics}). The median duration for the whole data set is 122 ms, which is close to every outliers median duration, except for outliers detected from the feature duration. Those same outliers have a probability of being slower than 250 ms of 92.4\%, which is about what is expected. The quantitative results concur with the qualitative results: DBSCAN works more efficiently with feature duration to detect outlying requests. Only those outliers are considered in the rest of this paper. For simplicity, the non-outlying requests -- the ones not detected as outliers by DBSCAN -- are called \textit{normal} requests even if they may contain anomalies.

As a short ablation study, DBSCAN was replaced with another outlier detection method called isolation forest. Similar results were obtained, and the same conclusions could be drawn. This indicates that it is not merely a preference of DBSCAN but rather the feature duration that is retaining useful information to detect outlying requests.

\subsection{Outlier Clustering  and Analysis}

K-means was applied to the feature counts of the outliers detected with DBSCAN. Two to ten clusters were studied. Using the elbow rule, three clusters were selected.

\begin{figure}[htp]
  \centering
  \includegraphics[width=.99\linewidth]{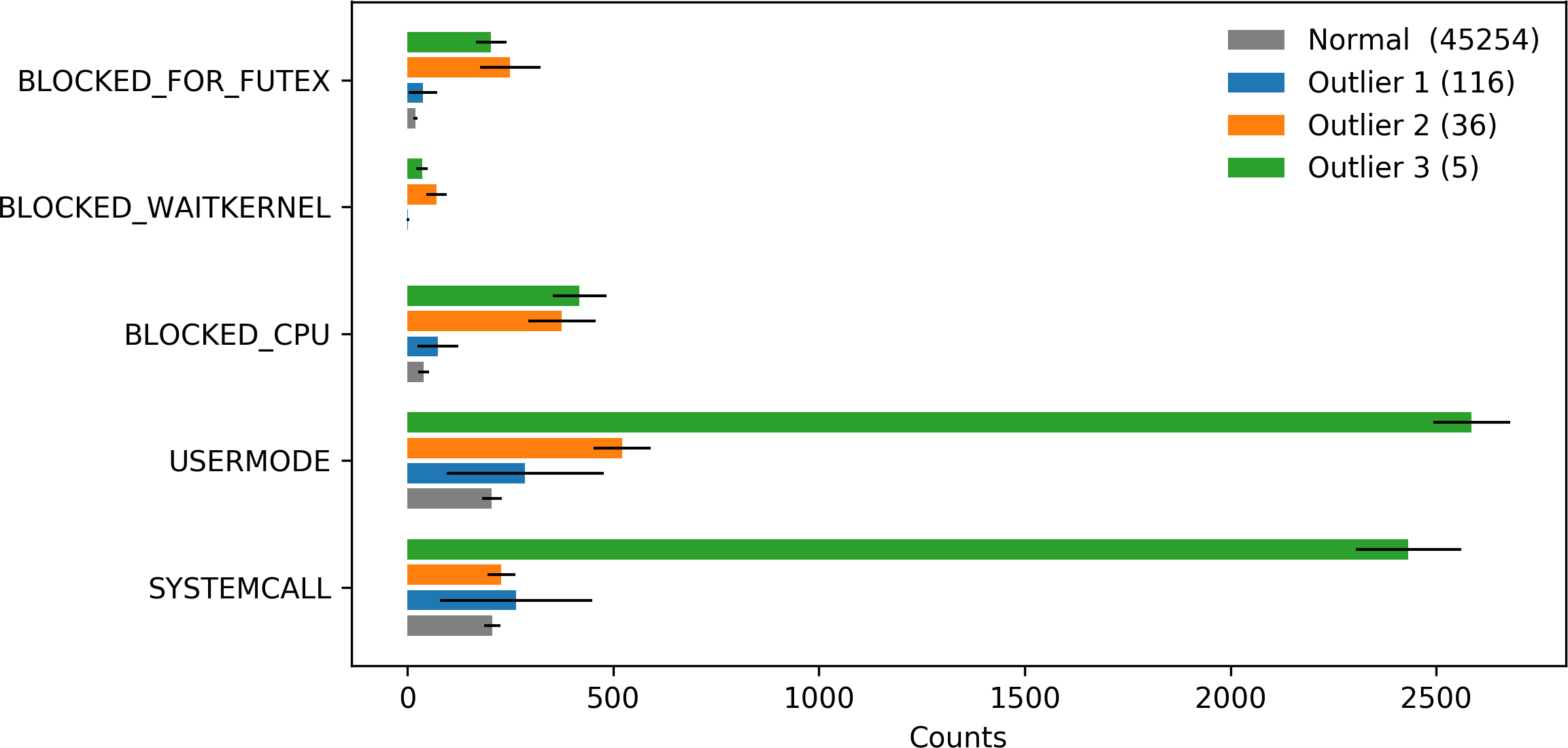}
  \caption{The average feature count in every cluster. The standard deviation is represented with a horizontal black line. There are three different types of outliers which depend on different resources.}
  \label{fig:kmeans_outliers}
\end{figure}

\begin{table}[htp]
\centering
\caption{Statistics of each cluster.}
\begin{tabular}{lrrrr}
 & Normal & Cluster 1 & Cluster 2 & Cluster 3 \\ \hline
Duration (ms) & 126 & 479 & \textbf{15512} & 559 \\
Number of system calls & 246 & 312 & 298 & \textbf{2690} \\
Distinct system calls & 26.0 & 26.4 & 26.1 & \textbf{29.0} \\ \hline
\end{tabular}
\label{tab:stat_clusters}
\end{table}

The clustering can be visualized by plotting the feature count histogram of each cluster and of normal requests (Fig. \ref{fig:kmeans_outliers}). In addition, Table \ref{tab:stat_clusters} compares statistics relating to requests in each group. Even though requests in the first cluster (blue) are nearly four times slower than the normal clusters, feature counts alone are not sufficient to set them apart. The high standard deviation of \texttt{usermode} counts indicates that this cluster is heterogeneous. The second cluster (orange) is composed of the slowest requests, which have a higher count than the normal ones, especially for \texttt{blocked} states. Finally, the last cluster (green) is characterized by an extremely high number of counts, especially for \texttt{usermode} and \texttt{systemcall} states. Although these are not the slowest requests, they have on average ten times more system calls and three more distinct system calls than the normal requests.

\begin{table}[htp]
\centering
\caption{Average unigrams, bigrams, and trigrams distinctive of each cluster.}
\setlength\tabcolsep{6pt}
\begin{tabular}{lrrrr}
Average n-gram & Normal & Cluster 1 & Cluster 2 & Cluster 3 \\ \hline
\texttt{write} & 1.0 & \textbf{2.3} & 1.1 & 1.6\\
\texttt{write write} & 0.0 & \textbf{1.3} & 0.1 & 0.6 \\
\texttt{write write write} & 0.0 & \textbf{0.9} & 0.1 & 0.4 \\ \hline
\texttt{cnnct} & 1.0 & 1.0 & \textbf{5.3} & 1.0\\
\texttt{cnnct cnnct} & 0.0 & 0.0 & \textbf{4.3} & 0.0 \\
\texttt{cnnct cnnct cnnct} & 0.0 & 0.0 & \textbf{3.7} & 0.0 \\ \hline
\texttt{fcntl} & 7.0 & 7.0 & 7.0 & \textbf{419.0}\\
\texttt{fcntl fcntl} & 2.0 & 2.0 & 2.0 & \textbf{305.0} \\
\texttt{fcntl fcntl fcntl} & 0.0 & 0.0 & 0.0 & \textbf{193.0} \\ \hline
\end{tabular}
\label{tab:n-grams}
\end{table}

Table \ref{tab:n-grams} shows differences in average n-grams between the normal requests and each cluster. They have been handpicked as they are unique to each cluster and explanatory of the underlying problem. Note that \texttt{connect} is shortened as \texttt{cnnct}. The complete list of n-grams is available on GitHub.

Requests in the first cluster differ from the normal ones only by their number of \texttt{write} system calls. It is not sufficient to draw any conclusion. One would need to apply domain-specific methods in order to determine the root cause or to show that those requests are false positives -- outliers that have normal behavior.

Requests in the second cluster are characterized by a high number of \texttt{connect} and its repetition. Note that if the flag \texttt{O\_NONBLOCK} is not set, \texttt{connect} is a blocking system call. This indicates that PHP cannot initiate a connection on a socket.

\begin{figure*}[!ht]
 \centering
    \includegraphics[width=.99\linewidth]{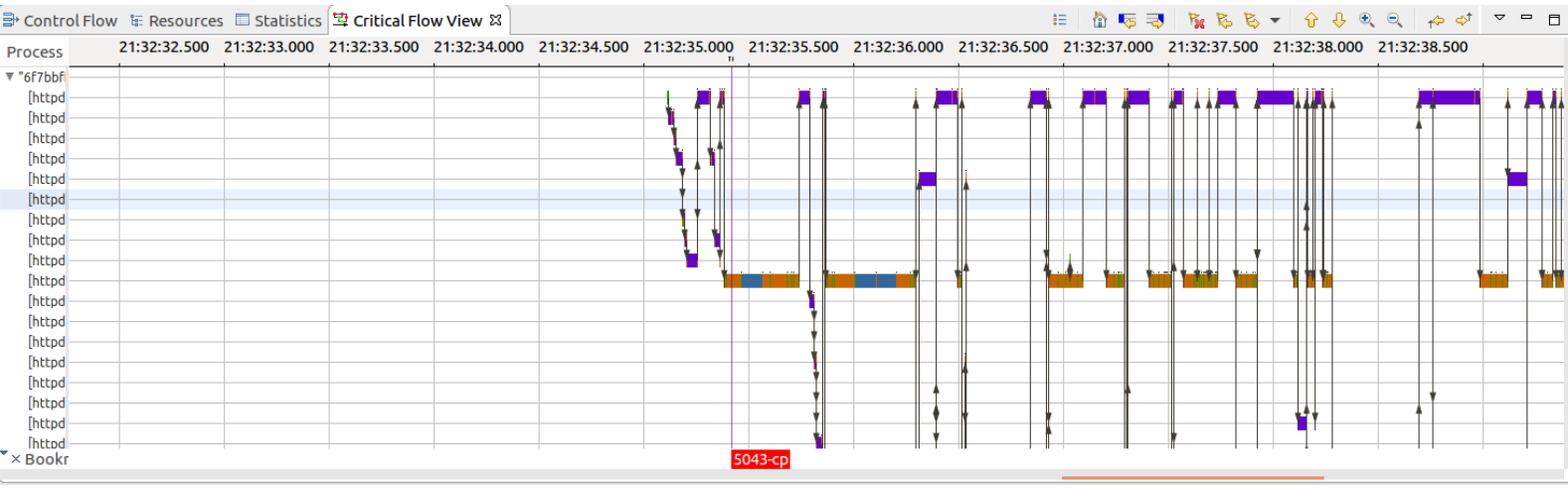}
    \caption{Critical Path analysis showing the contention between PHP processes}
    \label{criticalviewforall}
\end{figure*}

The third cluster is compelling as it explains cache issues. Further investigation into the PHP engine source code reveals that the \texttt{fcntl} system call is used for locking purposes. This, in fact, shows that there are unexpected blocking and waiting occurrences, hence some issues in the PHP processes while handling specific requests. At its core, PHP employs a cache mechanism called OPcache (OpCode Cache) to cache compiled script byte-codes in a shared memory for improved request handling performance. Whenever a script is compiled, the process checks the OPcache on the shared memory for an already-compiled code. If the code is missing, the PHP process compiles the code and writes the resulting byte-code to memory. However, when one process is writing into the shared memory, other processes must wait for the first one to release the lock before obtaining access and writing in the OPcache itself. This was the case for requests in the third cluster: PHP processes blocked each other to access the OPchache shared memory. A detailed analysis using a Critical Path View \cite{giraldeau} in Trace Compass (an open-source trace analysis tool\footnote{https://tracecompass.org}, confirms that there are issues among the PHP processes (shown in Figure \ref{criticalviewforall}) when concurrent requests are handled in the server.

\subsection{Visualization}

A qualitative display of the feature-duration space was obtained by reducing the dimensionality using isometric feature mapping (Isomap) \cite{tenenbaum_global_2000}. Isomap is a non-linear dimensionality reduction method that learns an embedding that preserves the intrinsic geometry of the data. This projection can be utilized to set a threshold duration, after which all requests are considered abnormal, speeding up outlier detection. As the cluster gathers requests up to 300 ms, it is a reasonable choice.

\begin{figure}[htp]
  \centering
  \includegraphics[width=.99\linewidth]{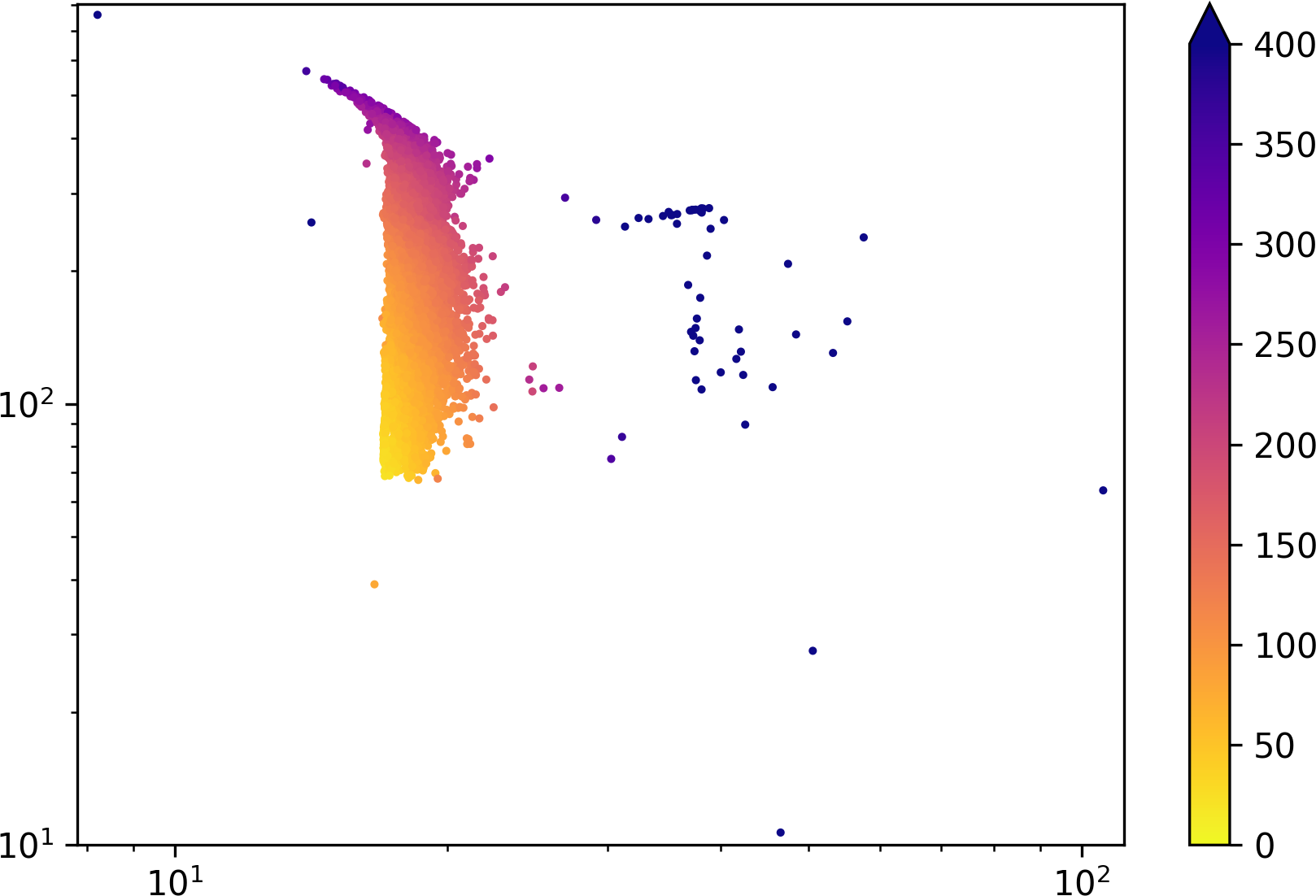}
  \caption{Visualization of the feature-duration space using ISOMAP. The color indicates the duration of the request in milliseconds. There are about 10 clear outliers outside of the window. Note the log scale on both axes.}
  \label{fig:visualization_duration}
\end{figure}

\subsection{Computational Cost}

The proposed method for clustering normal and abnormal requests, based on kernel-level features from thread interaction in the critical path, will prove useful only if it can be efficiently scaled.

Experiments were conducted on a modern computer equipped with a 6-core processor and 32 GB of main memory. Requests were generated using a Wordpress website running on Linux 18.04 with PHP 7.0.32 and Apache 2.4.18. The trace was collected using LTTng 2.11.

\subsubsection{Tracing Cost}

Experiments were performed using between 1 and 1,000 clients with the following tracing configurations:
\begin{itemize}
    \item No tracing.
    \item Minimal tracing: only a small subset of events required for analysis is enabled.
    \item Full tracing: all kernel space and user space events are enabled.
\end{itemize}

\begin{figure}[htp]
  \centering
  \includegraphics[width=.99\linewidth]{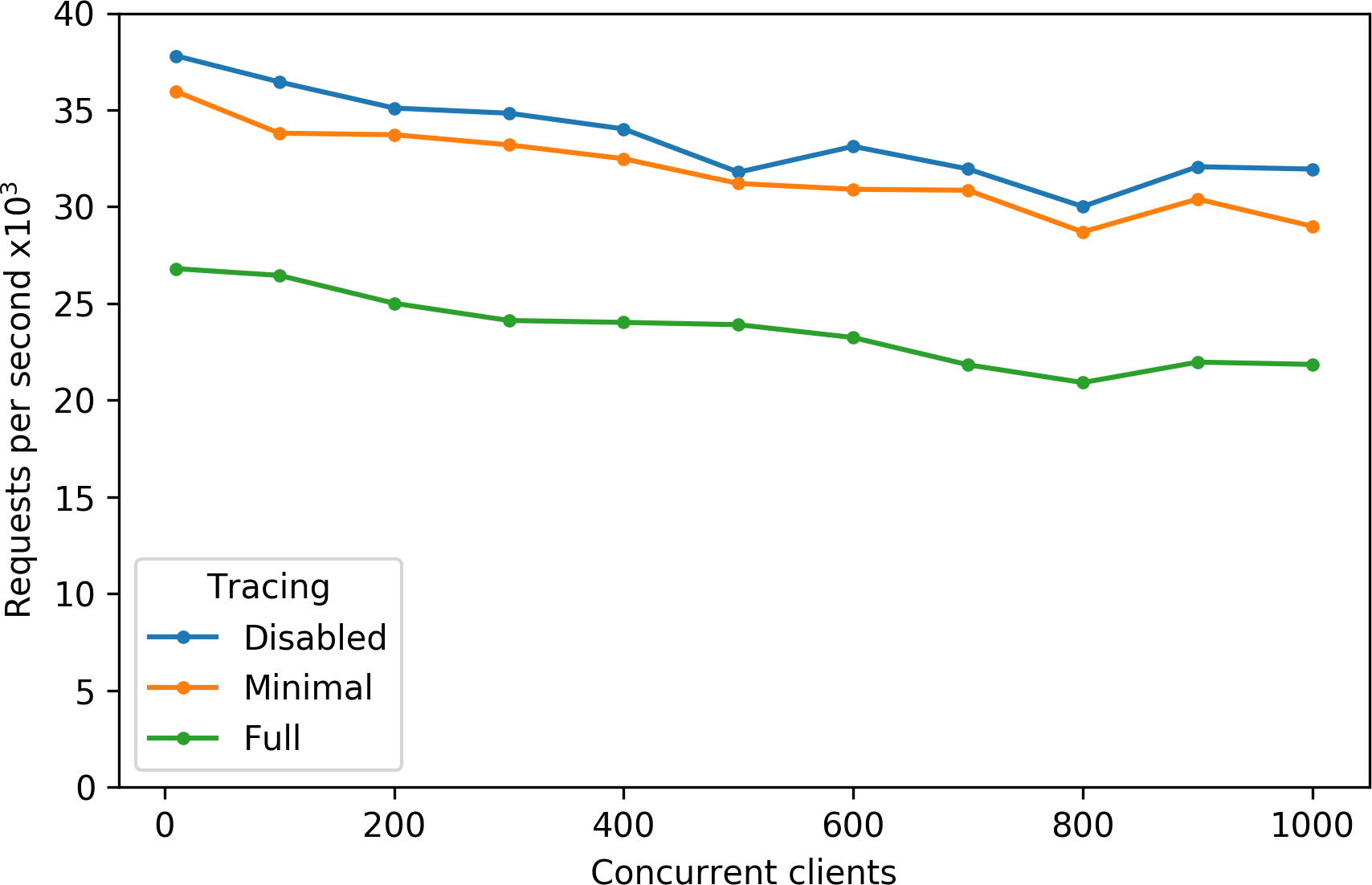}
  \caption{Cost of the different tracing configurations, measured by the number of request per second.}
  \label{tracingcosts}
\end{figure}

The capacity of the server to handle requests per second is shown in Figure \ref{tracingcosts}, using the above tracing configurations. When all kernel space and user space events are enabled, tracing has a significant impact on the performance, with a slowdown of $29.6\pm2.0\%$. This is mostly due to the kernel tracing, since it collects and stores a large amount of system execution details to the disk. However, the proposed analysis does not require enabling all the tracepoints. In user space (i.e., PHP core), only two events are enabled: $request\_start$ and $request\_end$. In kernel space, only scheduling, system call, timer, and waking up events are necessary to extract all the essential execution states. With this minimal setup, the overhead is reduced to $5.1\pm1.9\%$, coming mostly from the kernel-enabled events.

\subsubsection{Analysis Cost}

Note that the analysis is written in Python and is not parallelized unless explicitly specified. Much faster run times could be achieved with a C or C++ implementation.

\begin{table}[htp]
    \centering
    \caption{Average run time of each step plus or minus the standard deviation. Each step was run ten times.}
    \begin{tabular}{lr}
        Step & Run time \\\hline
        Feature extraction & 29.31 $\pm$ 1.08s\\
        Outlier detection & 5.69 $\pm$ 0.83s\\
        Outlier clustering & 0.04 $\pm$ 0.00s\\
        N-gram analysis & 17.59 $\pm$ 0.16s\\\hline
    \end{tabular}
    \label{tab:perf}
\end{table}

Creating the feature bag-of-word took 1.17 $\pm$ 0.12s, as it is linear in the number of examples. It is not included in Table \ref{tab:perf} since the feature duration was used instead. Furthermore, as the bag-of-word matrix is often sparse, it can be efficiently stored in memory.

The run time complexity of DBSCAN is $\mathcal{O}(n \text{log} n)$, assuming it uses an indexing structure and an $\epsilon$ which gives $\text{log} n$ neighbors an average \cite{Ester:1996:DAD:3001460.3001507}. The parallelized scikit-learn implementation of DBSCAN was used.

K-means is known to be an efficient method, with an average complexity of $\mathcal{O}(knT)$ where $k$ is the number of neighbors and $T$ is the number of iterations. The parallelized scikit-learn implementation of k-means was also used.

Even though n-gram is merely counting, and there are a limited number of possibilities (31 unigrams, 135 bigrams, and 334 trigrams in detected outliers), this is the slowest step as the results were stored in a data frame before being displayed and saved. This makes the data manipulation easier, although it is not necessary and can be optimized for production.

\section{Conclusion and Future Work}
\label{conclusion}

In this paper, we introduced a pipeline for the automatic detection of performance issues in web requests. This includes both user space and kernel space feature extraction methods, which proved to be useful for detecting outliers and clustering them in a relevant manner. To answer our first research question: it is indeed possible to automatically and efficiently detect anomalies from trace data. A simple statistical analysis was able to guide us to the root cause of two different anomalies. This indicates that comparing run time behavior gives meaningful insights into the root cause.

As future work, we would like to compare our method with ones specifically designed for sequences, even if some assumptions are broken. Notably, Hidden Markov Models and SEQDBSCAN -- an extension of DBSCAN for sequences -- seem promising. We would also like to extend the cluster analysis to context calling trees (CCT) or enhanced context calling trees (ECCT) \cite{7469402} in order to determine if the first cluster in Figure \ref{fig:kmeans_outliers} and Table \ref{tab:stat_clusters} is a group of normal requests with a different behavior, or merely a group of unknown anomalies.

\section{ACKNOWLEDGMENT}

We would like to gratefully acknowledge the Natural Sciences and Engineering Research Council of Canada (NSERC), Ericsson, Ciena, and EffciOS for funding this project.


\end{document}